\documentclass[letterpaper, 10 pt, conference]{ieeeconf}  % Comment this line out if you need a4paper

\IEEEoverridecommandlockouts                              % This command is only needed if                     % you want to use the \thanks command

\usepackage{cite}
\usepackage{mathtools}
\usepackage{amsmath,amssymb,amsfonts}
\usepackage[english]{babel}
\newtheorem{definition}{Definition}[section]

\newtheorem{lemma}[definition]{Lemma}

\newtheorem{thm}{Theorem}[section]

\newtheorem{remark}{Remark}

\usepackage{algorithmic,balance}
\usepackage{graphicx}
\usepackage{textcomp}
\usepackage{xcolor}
\usepackage{booktabs}
\usepackage{multirow}
\usepackage{hyperref}
\usepackage{subfig}

\title{\LARGE \bf
A Scalable Approach for Safe and Robust Learning via Lipschitz-Constrained Networks

\thanks{$^{1}$The Bradley Department of Electrical and Computer Engineering, Virginia Tech, Blacksburg, VA, USA. 
        {\tt\small Emails: \{zabdeen, jinming\}@vt.edu}\\
        $^{2}$Elmore Family School of Electrical and Computer Engineering, Purdue University, West Lafayette, IN, USA.
        {\tt\small Emails: kekatos@purdue.edu}} 
        
       }%

\author{Zain ul Abdeen$^{1}$, Vassilis Kekatos$^{2}$, Ming Jin$^{1}$ % <-this % stops a space
}

\begin{document}

\maketitle
\thispagestyle{empty}
\pagestyle{empty}

%%%%%%%%%%%%%%%%%%%%%%%%%%%%%%%%%%%%%%%%%%%%%%%%%%%%%%%%%%%%%%%%%%%%%%%%%%%%%%%%
\begin{abstract}
Certified robustness is a critical property for deploying neural networks (NN) in safety-critical applications. A principle approach to achieving such guarantees is to constrain the global Lipschitz constant of the network. However, accurate methods for Lipschitz-constrained training often suffer from non-convex formulations and poor scalability due to reliance on global semidefinite programs (SDPs). In this letter, we propose a convex training framework that enforces global Lipschitz constraints via semidefinite relaxation. By reparameterizing the NN using loop transformation, we derive a convex admissibility condition that enables tractable and certifiable training.
% This letter addresses the problem of training NN under global Lipschitz constraints to enhance certifiable robustness. We propose a semidefinite convexification framework that reparametrize the network via loop transformation, enabling convex admissibility condition tractable for training. 
While the resulting formulation guarantees robustness, its scalability is limited by the size of global SDP. To overcome this, we develop a \textit{randomized subspace linear matrix inequalities (RS-LMI)} approach that decomposes the global constraints into sketched layerwise constraints projected onto low-dimensional subspaces, yielding a smooth and memory-efficient training objective. Empirical results on MNIST, CIFAR-10, and ImageNet demonstrate that the proposed framework achieves competitive accuracy with significantly improved Lipschitz bounds and runtime performance. 
\end{abstract}

%%%%%%%%%%%%%%%%%%%%%%%%%%%%%%%%%%%%%%%%%%%%%%%%%%%%%%%%%%%%%%%%%%%%%%%%%%%%%%%%
	\section{Introduction}
    Deep learning (DL) has achieved success in many areas, such as computer vision, natural language processing, and reinforcement learning \cite{fazlyab2019efficient,goodfellow2016deep,sutskever2014sequence}. However, its adaption in safety-critical systems remains limited due to its lack of interpretability, fragility, and absence of robustness guarantees~\cite{leino2021globally,stoica}.
% The advances in deep learning (DL) have impacted many areas, such as computer vision, natural language processing and as well as in reinforcement learning \cite{goodfellow2016deep,sutskever2014sequence}.However, the use of DL for safety-critical tasks in the real world is challenged by its opaqueness, fragility, and vulnerability~\cite{krizhevsky2012imagenet,stoica}. 
For example, imperceptible adversarial perturbations can cause models to misclassify inputs with high confidence~\cite{szegedy2013intriguing}. This has spurred research into improving NN robustness.
% Adversarial attack is a problem that has recently been tackled increasingly in number of different ways 
Several defense strategies including adversarial training \cite{wang2023direct} and defensive distillation\cite{papernot2016distillation},
% have been proposed. While these techniques 
improve robustness under specific threat models, but they do not offer formal guarantees and often generalize poorly to unseen attacks. 
% to improve robustness, yet neither of which providing formal guarantees. The adversarial trained network is robust to specific attacks seen during training, but can still be vulnerable to unseen attacks. In other words, the network is robust against norm-bounded adversarial perturbation \cite{liu2024towards,wong2018provable}. Several works pursue this by optimizing worst-case margins or loss bounds e.g., maximizing margins during training \cite{huang2021training,fazlyab2023certified}. While the aforementioned robustness measure are restricted to classification problems and as they only provide guarantees in uncertainty sets during training data points.
A promising alternative is \textit{certifiable robustness}, where models are trained with guarantees that their outputs remain consistent under norm-bounded input perturbations~\cite{liu2024towards}. Many such methods optimize margin-based loss bounds~\cite{fazlyab2023certified},
% or adversarial losses during training
to yielf local robustness certificates, which are often restricted to classification tasks and specific threat models. 
% Unlike certified robust training, 
In contrast, we quantify robustness by explicitly bound the global sensitivity of NN to input perturbation using the Lipschitz constant $L$,
% we bound the sensitivity of the function to input perturbation by bounding the Lipschitz constant $L$ to quantify the robustness of NN~\cite{pauli2021training}, 
i.e., for any function $\Psi:\mathbb{R}^{n_{x}}\to\mathbb{R}^{n_{y}}$, if the input changes from $u$ to $w$ the Lipschitz constant gives an upper bound on how much the output changes, formally defined as~\cite{pauli2021training}:
\begin{equation}
    \|\Psi(u)-\Psi(w)\|_2\leq L\|u-w\|_2\quad \forall u,w\in \mathbb{R}^{n_{x}}
\end{equation}
A smaller $L$ implies a tighter worst-case bound on output deviation and thus greater robustness.  
Tight bounds on $L$ support closed-loop stability analysis \cite{brunke2022safe}, adversarial certificates \cite{pauli2021training}, and generalization guarantees \cite{bartlett2017spectrally}.
%%%%%%%%%%%%%%%
% A smaller $L$ 
% % indicates lower sensitivity to input perturbation, which 
% implies greater robustness to worst-case perturbation. 
% % A smaller $L$ indicates lower sensitivity to input perturbation implies greater resistance to worst-case perturbation, equivalent to high robustness. 
% When~$\Psi$ is characterized by DNN, tight bounds on $L$
% % its Lipschitz constant 
% can play an important role in analyzing closed-loop stability~\cite{brunke2022safe}, certifying adversarial robustness~\cite{pauli2021training}, and deriving generalization guarantees~\cite{bartlett2017spectrally}.
%%%%%%%%%%%%%%
% the stability analysis of the closed-loop system \cite{brunke2022safe}, the robustness certificate of the NN-based classifier against adversarial attacks \cite{pauli2021training}, and in derivative of generalization bounds \cite{bartlett2017spectrally}. 
 
% while yet another one is to use Lipschitz constants as a robustness measure that indicate the sensitivity of the output to perturbations in the input\cite{fazlyab2019efficient}. 

Computing the exact Lipschitz value is NP-hard even for shallow ReLU networks~\cite{virmaux2018lipschitz}, leading to approximation methods. A common heuristic is to upper-bound the Lipschitz constant via product of spectral norms of the weight matrices~\cite{szegedy2013intriguing}. While this approach is computationally efficient and forms the basis for Lipschitz regularization during training~\cite{gouk2021regularisation,leino2021globally}, but often yields loose bounds. A more accurate approach, known as LipSDP~\cite{fazlyab2019efficient}, formulates tight upper bounds using SDP. Although theoretically compelling, 
% the SDP formulation is computationally expensive and memory-intensive. Moreover,
LipSDP is generally nonconvex in network parameters, making it incompatible with standard gradient-based training. To address this, Pauli et al.~\cite{pauli2021training}, incorporated LipSDP constraints into training,
% to encourage robustness,
but this requires conservative slope bounds on activation functions and scales poorly beyond small networks.
These limitations highlight two key challenges in Lipschitz-constrained learning: (i) the \emph{nonconvexity} of the learning constraints with respect to the network parameters; and (ii) the \emph{scalability} of training procedures with tight robustness guarantees.
\textbf{Contributions:}
We propose a scalable and memory-efficient framework for certifiably robust neural network training that addresses both challenges. Specifically, first, we propose \emph{Lip-Loop}, a convex reparameterization of the LipSDP~\cite{fazlyab2019efficient} constraint using loop transformation, enabling LMIs to be enforced directly during training via SDP. 
% a reparameterization scheme based on loop transformation that reformulates non-convex constraints in LipSDP \cite{fazlyab2019efficient} into convex LMIs, which can be enforced directly during training via SDP
% allowing efficient integration into training pipeline.
Second, to address scalability, we propose \emph{RS-LMI} method that projects constraints onto low-dimensional Gaussian subspaces, yielding per-layer certificates via eigenvalue-based matrix inequalities. This yields a differentiable, memory-efficient objective compatible with first-order optimization methods. We evaluate both approaches across real-world benchmarks, showing that they achieve competitive accuracy with significantly improved Lipschitz bounds and computational efficiency.
% By composing per-layer certificates, \textit{RS-LMI} provides global Lipschitz upper bound with high-probability guarantees. The combined approach enables deep network training with certified robustness and significantly reduced cost. 

% method that certifies the Lipschitz bound of each network layer by projecting its weight operator onto low-dimensional Gaussian subspaces. This yields matrix inequalities that are efficiently converted into differentiable penalties using eigenvalue-based projections, allowing compatibility with first-order optimization methods.
% % and enforcing matrix inequalities. The resulting constraints are transformed into differentiable penalties using eigenvalue-based projections, facilitating compatibility with first-order optimization. 
% By composing per-layer subspace certificates, RS-LMI yields a global Lipschitz bound with high-probability guarantees. The combined approach 
% % enables training of deep networks with training of deep networks, with tighter robustness guarantees and reduced computational cost compared to existing methods.
% enables training of deep networks with certified robustness and significantly reduced computational cost.

    \textbf{Related work:} 
    % The Lipschitz constant play a central role in many works on training a certifiably robust neural network, especially for $\ell_2$ norm robustness. Since naturally trained networks usually have very large global Lipschitz bounds \cite{szegedy2013intriguing}. 
    % A variety of approaches have explored the estimation and certifications of Lipschitz constants in the context of NNs verification
    Numerous works have studied Lipschitz constant estimation for neural network verification~\cite{qin2019verification, shi2022efficiently,fazlyab2023certified,chord3}. However, the associated constraints are nonconvex in the network parameters.
    % , limiting their integration during training. 
    Verification through convex relaxations has been explored in~\cite{qin2019verification}, aligning with our goal of general robustness certification beyond classification. Our first work is inspired by \cite{yin2021imitation,ul2022learning}, where constraint convexification was employed using loop transformation to synthesize NN controller with stability and safety grantees. SDP-based training methods~\cite{pauli2021training,pauli2022neural} have been demonstrated on small networks but face scalability limitations. Chordal decomposition ~\cite{chord3,zheng2021chordal} has been used to break down large SDPs into smaller subproblems, but the resulting formulations still inherit nonconvexity with respect to network parameters. 
    Recently in~\cite{wang2023direct} employs sparse LipSDP structures that enable learning through direct scalable parameterization, but this requires careful design of the network architecture with prescribed Lipschitz
guarantees, leading to non-trivial certified robust accuracy.
    % Recently, \cite{wang2023direct} use the sparse structure of SDP condition for DNNs, leading to a scalable direct parameterization of the SDP-based Lipschitz condition, that enable learning of lipschitz bounded network via standard gradient methods, avoiding the complex projection steps or barrier function computation. But this allows the design of specific network structures with prescribed Lipschitz guarantees, leading to non-trivial certified robust accuracy.
Our second approach is inspired by recent advances in optimization for structured SDPs~\cite{zheng2021chordal,garstka2021cosmo}, particularly methods that reduce SDP feasibility to eigenvalue problems, solvable with subgradient descent~\cite{wang2024scalability}. We build on these insights to enable certifiable and efficient training of Lipschitz-constrained deep networks.

\textbf{Notation:}Let $\mathbb{S}^{n}$, $\mathbb{S}^{n}_{+}$, $\mathbb{S}^{n}_{++}$ denote the sets of $n \times n$ symmetric, positive semi-definite and positive definite matrices, respectively. For any matrix 
$A \in \mathbb{S}^{n}$, the inequality  $A \succeq 0 $ and $A \succ 0$ indicates positive semi-definiteness and positive definiteness, respectively.

	%%%%%%%%%%%%%%%%%%%%%%%%%%%%%%%%%%%%%%%%%%%%%%%%%%%%%%%%%%%
	% \section{Problem Formulation}
	% \subsection{Problem Statement}
	% We consider specifications on outputs in relation to inputs that vary across instances. Formally, we define a multi-layer feed-forward neural network (NN) mapping $\Psi(\cdot;\theta):\mathcal{X}\to \mathcal{Y}$ parameterized by a weight vector $\theta \in \mathbb{R}^{n_\theta}$. Sets $\mathcal{X}\subseteq \mathbb{R}^{n_x}$ and $\mathcal{Y}\subseteq \mathbb{R}^{n_y}$ are respectively the sets wherein NN inputs and outputs can lie. The main goal is to learn a NN parameters $\theta$ with certified Lipschitz bound of $L$, i.e.,
\section{Problem Formulation}
	\subsection{Problem Statement}
We consider the problem of training a multi-layer feed-forward NN, defined as $\Psi(\cdot;\theta):\mathcal{X}\to \mathcal{Y}$ parameterized by a weight vector $\theta \in \mathbb{R}^{n_\theta}$. Here, $\mathcal{X}\subseteq \mathbb{R}^{n_x}$ and $\mathcal{Y}\subseteq \mathbb{R}^{n_y}$ are respectively the input-output sets. Our goal is to identify parameters $\theta$ that minimize a task-specific loss~$\mathcal{L}(\Psi_{\theta})$, while satisfying the global Lipschitz constraint, i.e.,
% with certified Lipschitz bound of $L>0$, i.e.,

    % We also define an $m$-way specification $F:\mathcal{X}\times \mathcal{Y}\to \mathbb{R}^{m}$, and its associated specification  set $\mathcal{S}(x)\coloneqq \{y\in\mathbb{R}^{n_y}: F(x,y)\leq 0\}$. Our aim is to find a  parameter $\theta$ such that 
	% \begin{equation}\label{eq1}
	% \Psi(x; \theta)\in \mathcal{S}(x),~~\forall x\in \mathcal{X}.
	% \end{equation}
    \begin{equation}\label{eq1}
	\min_{\theta}\mathcal{L}(\Psi_{\theta})\quad \text{s.t.}\quad \text{Lip}(\Psi_{\theta})\leq L
	\end{equation}
     To simplify notation, we henceforth omit the dependence of $\Psi$ on $\theta$. Above formulation \eqref{eq1} subsumes several robustness-constrained learning settings,
     % The above formulation subsumes a family of problems in machine learning and control
     such as imposing fairness, adversarial robustness, and reinforcement learning~\cite{barbara2024robust,fazlyab2020safety,qin2019verification}. 
     Note that the admissible parameter set $\Theta \coloneqq \{\theta \in \mathbb{R}^{n_\theta}: \eqref{eq1} \  \text{is satisfied}\}$ is generally nonconvex due to the nonlinearity inherent in $\Psi$ and the constraints $\text{Lip}(\Psi_{\theta})\leq L$.  Thus, optimizing directly over the nonconvex set $\Theta$ is intractable, 
     % Granted that searching within the nonconvex admissible set $\Theta$ is intractable,
     % our strategy is to compute a convex inner approximation of $\Theta$. 
     our strategy is to propose a convex inner approximation captured through a semidefinite program. To this end, we first reformulate the input-output mapping of an NN to isolate the nonlinearities of activation functions.

\subsection{Abstracting Non-Linearities of NN} 
	% \subsection{Isolating NN nonlinearities}
	The input-output mapping of a multi-layer NN with $l$ layers can be described by the recursive equations:
	\begin{equation}\label{nn1}
	\centering
	\begin{split}
	x^0&=x\\ x^{k+1}&=\phi^k(W^kx^k+b^k),~~~~~k=0,1,\dots,l-1\\ \Psi(x)&=W^lx^l+b^l
	\end{split}
	\end{equation}
	where $x\in \mathcal{X}$ is the input, $W^k \in \mathbb{R}^{n_{k+1}\times n_k}$ and $b^k\in \mathbb{R}^{n_{k+1}}$ are the weight matrix and bias vector of the $(k+1)\text{-th}$ layer, respectively; and  $n=\sum_{k=1}^{l}n_k$ is the number of neurons. The mapping~$\phi^k$ applies a nonlinear scalar activation function $\psi:\mathbb{R}\to \mathbb{R}$ on each one of the entries of its vector argument $W^kx^k+b^k$. The mapping can be defined as:
	\begin{eqnarray}
	\phi^k(x)=[\psi(x_1),\psi(x_2),\dots, \psi(x_{n_k})]^{\top}.
	\end{eqnarray}
	Typical choices for 
    % the scalar activation function 
    $\psi$ include the $\tanh$, the sigmoid, and the rectified linear unit (ReLU). 
    % To facilitate subsequent derivations, 
    To facilitate the convex analysis, we isolate the non-linear and linear components of an NN as in \cite{fazlyab2020safety}, 
    allowing us to impose global Lipschitz constraints through semidefinite relaxation. Let $v^k=W^k x^k +b^k$ denote the input to the activation function in layer $k+1$. Then, the NN defined in~\eqref{nn1} can be rewritten as
	\begin{align}\label{A}
	\begin{bmatrix} v_\phi\\ \Psi(x)  \end{bmatrix}
	&=
	N\begin{bmatrix} x \\ x_\phi  \end{bmatrix}+ \begin{bmatrix} b_\phi \\ b^l\  \end{bmatrix}\\
	{x_\phi}&=\phi (v_\phi)
	\end{align}
where,
$\tiny{v_\phi = \begin{bmatrix} v^0 \\ \vdots \\ v^{l-1} \end{bmatrix}, 
% x_\phi = \begin{bmatrix} x^1 \\ \vdots \\ x^{l} \end{bmatrix}
 b_\phi = \begin{bmatrix} b^0 \\ \vdots \\ b^{l-1} \end{bmatrix}, \phi(v_\phi) = \begin{bmatrix} \phi^0(v^0) \\ \vdots \\ \phi^{l-1}(v^{l-1})\end{bmatrix}}.$ \color{black}
The matrix $N$ encodes the linear operations across layers. 
% depends on the NN weights 
% and can be partitioned as follows:
% \begin{align}\nonumber
%  N&= \small{\left[ 
% \begin{array}{c | c}
% \begin{array}{c}
%      W^{0}  \\
%      0\\
%      \vdots
% \end{array}&
% \begin{array}{c c c c c} 
% 		 0 & \dots&0&0&0\\ 
% 		 W^{1} & \dots& 0&0&0\\ 
% 		\vdots &\dots& \vdots & W^{l-1}&0 
% 	\end{array} \\	 
% 	\hline 
% 	\begin{array}{c}
% 		0
% 	\end{array}&
% 	\begin{array}{ccccc}
% 	~~~~0&\dots&0&~~~0&~~~~W^{l}
% 	\end{array}
% \end{array} 
% \right]} =\begin{bmatrix}N_{vx} & N_{vx_{\phi}} \\ N_{\Psi x} & N_{\Psi x_{\phi}}  \end{bmatrix}. 
% \end{align}
\begin{align}\nonumber
N &= 
\left[
\begin{array}{c|cccc}
W^0 & 0& \cdots & 0              & 0 \\
0       & W^1     & \cdots & 0              & 0 \\
\vdots  & \vdots  & \ddots & \vdots  & \vdots   \\
0       & 0       & \cdots & W^{l-1} & 0        \\
\hline
0       & 0       & \cdots & 0             & W^l
\end{array}
\right] 
=
\begin{bmatrix}
N_{vx} & N_{vx_{\phi}} \\
N_{\Psi x} & N_{\Psi x_{\phi}}
\end{bmatrix}
\end{align}

    % A key challenge in analyzing NN is the composition of nonlinear activation functions.
    
A key challenge in analyzing NN is the composition of non-linear activation functions.
    Most activation functions has common feature of bounded derivatives. Technically, a function $\psi$ is sloped-restricted in the interval $[\alpha,\beta]$ if 
\begin{equation}\label{sloperestricted}
        { \alpha \leq \frac{\psi(x_j)-\psi(x_{i})}{x_{j}-x_{i}}\leq \beta,\quad \forall x_{i},x_{j} \in \mathbb{R}}.
     \end{equation}
% By exploiting the common pattern of activation functions, a viable approach is to employ slope bounded.
%  \begin{definition}[\cite{fazlyab2020safety}]
%      Given $\alpha\leq \beta$, function $\psi$ lies in slope $[\alpha,\beta]$ if 
% \begin{equation}\label{sloperestricted}
%         { \alpha \leq \frac{\psi(x_j)-\psi(x_{i})}{x_{j}-x_{i}}\leq \beta,\quad \forall x_{i},x_{j} \in \mathbb{R}}
%      \end{equation}
%  \end{definition}
 Thanks to the property of bounded slope in \eqref{sloperestricted}, we can provide an inner convex approximation of the nonlinear behavior of the activation function. 
 % The slope restricted constraints \eqref{sloperestricted} enables us to derive a incremental quadratic constraints (IQCs), which are useful for characterizing nonlinear behavior in convex form. 
 The following lemma formalizes the relationship for a vector-valued function $\phi:\mathbb{R}^{n_{\phi}}\to \mathbb{R}^{n_{\phi}}$, where $n_{\phi}=\sum_{k=0}^{l-1}n_{k+1}$ and local bounds can be concatenated in the form of vectors $\alpha_{\phi},\beta_{\phi}\in \mathbb{R}^{n_{\phi}}$. 
 
 % For every neuron the slope restricted property \eqref{sloperestricted} can be written as an incremental quadratic constraints (IQC). 
 
 \begin{lemma}[\cite{fazlyab2020safety}]  Suppose $\phi$ is slope bounded by $[\alpha_{\phi},\beta_{\phi}]$ coordinate-wise for all $x,y$. Define the set $\mathcal{T}_{n_{\phi}}:=\{T= \text{diag}(\lambda_{1},\dots,\lambda_{n_{\phi}})\in \mathbb{S}^{n_{\phi}}_{+}:\lambda_{i}\geq 0\}$.
 % for multiplier $ \lambda_{ij}\geq 0$
 Then for any~$T\in \mathcal{T}_{n_{\phi}}$, the function $\phi$ satisfies incremental quadratic constraints (IQCs) as
 % $\phi^k(x)=[\psi(x_1),\psi(x_2),\dots, \psi(x_{n_k})]^{\top}$ satisfies 
% \begin{gather}\label{QCac}
% 		\small{\begin{bmatrix} x -y \\  \phi(x)-\phi(y)  \end{bmatrix}^\top
% 		\begin{bmatrix}
% 		-2\alpha_{\phi}\beta_{\phi} T & (\alpha_{\phi}+\beta_{\phi})T \\ (\alpha_{\phi}+\beta_{\phi})T & -2T
% 		\end{bmatrix}
% 		\begin{bmatrix} x-y  \\  \phi(x)-\phi(y)  \end{bmatrix}}
% 		\geq 0
% 	\end{gather} 
\begin{gather}\label{QCac}
\scalebox{0.88}{$
\begin{bmatrix} x - y \\ \phi(x) - \phi(y) \end{bmatrix}^\top
\begin{bmatrix}
-2\alpha_{\phi}\beta_{\phi} T & (\alpha_{\phi} + \beta_{\phi})T \\
(\alpha_{\phi} + \beta_{\phi})T & -2T
\end{bmatrix}
\begin{bmatrix} x - y \\ \phi(x) - \phi(y) \end{bmatrix}
\geq 0
$}
\end{gather}     
 \end{lemma}

  \section{Admissibility analysis for Lipschitz Constant Estimate}
 Based on slope restricted IQC abstractions of nonlinear activations and $\mathcal{S}$-procedure \cite{beta}, the following result provides the admissibility condition of a fixed NN for the estimation of Lipschitz bound. The resulting method provides tighter estimates than norm-product heuristic.
 % In the following, we outline a method to estimate bounds on the Lipschitz constant of multi-layer NNs exploiting the slope-restricted structure of the nonlinear activation functions,as it was shown in \cite{fazlyab2019efficient}. This method named LipSDP yieldsmore accurate bounds than trivial bounds, i.e., the product of the spectral norms of the weights.

% \noindent \textbf{\textcolor{red}{convex} in variables} $Q_{1}=T^{-1}$, $K_{1}=\tilde{N}_{vx},K_{2}=\tilde{N}_{vx_{1}}Q_{1}, K_{3}=\tilde{N}_{\Psi x}, K_{4}=\tilde{N}_{\Psi x_{1}}Q_{1}$.

\begin{thm}[\text{LipSDP} \cite{fazlyab2019efficient}]\label{lipsdpthm}
Suppose $\phi$ is slope bounded by $[\alpha_{\phi},\beta_{\phi}]$. There exists $L>0$, and $T\in\mathcal{T}_n$ such that:
\begin{align}\label{eq4}
\begin{split}
     \begin{bmatrix} N_{vx} & N_{vx_{\phi}} \\ 0 &I_{n_{\phi}} \end{bmatrix}^{\top} &
  \begin{bmatrix} -2\alpha_{\phi}\beta_{\phi} T & (\alpha_{\phi} +\beta_{\phi})T \\ (\alpha_{\phi}+\beta_{\phi})T & -2T \end{bmatrix}  \begin{bmatrix} N_{vx} & N_{vx_{\phi}} \\ 0 &I_{n_{\phi}} \end{bmatrix}\\&+
		\begin{bmatrix} N_{\Psi x}^{\top}N_{\Psi x}-L^{2}I & N_{\Psi x}^{\top}N_{\Psi x_{\phi}} \\ N_{\Psi x_{\phi}}^{\top}N_{\Psi x} & N_{\Psi x_{\phi}}^{\top}N_{\Psi x_{\phi}} \end{bmatrix}  \preceq 0
\end{split}
  \end{align}
%   \begin{align}\label{eq4}
% \begin{split}
%      \begin{bmatrix} N_{vx} & N_{vx_{\phi}} \\ 0 &I_{n_{\phi}} \end{bmatrix}^{\top} &
%   \begin{bmatrix} -2\alpha\beta T & (\alpha +\beta)T \\ (\alpha+\beta)T & -2T \end{bmatrix}  \begin{bmatrix} N_{vx} & N_{vx_{\phi}} \\ 0 &I_{n_{\phi}} \end{bmatrix}\\&+
% 		\begin{bmatrix} -L^{2}I & 0 & 0 \\ 0 & 0 & 0 \\ 0 & 0 &  W_{l}^{\top}W_{l} \end{bmatrix}  \preceq 0
% \end{split}
%   \end{align}
  Then, $\Psi$ is globally Lipschitz continuous with bound $L$. 
%   where
% \begin{align*}
% A = \begin{bmatrix}
% W_0     & 0       & \cdots & 0      & 0 \\
% 0       & W_1     & \cdots & 0      & 0 \\
% \vdots  & \vdots  & \ddots & \vdots & \vdots \\
% 0       & 0       & \cdots & W_{l-2}& 0 
% \end{bmatrix},\quad
% B = \begin{bmatrix}
% 0 & I_n
% \end{bmatrix}
% \end{align*}
% and $M = \sum_{i=1}^{n} \lambda_i \mathbf{e}_i \mathbf{e}_i^\top + \sum_{\substack{1 \le i < j \\ \le n}} \lambda_{ij} (\mathbf{e}_i - \mathbf{e}_j)(\mathbf{e}_i - \mathbf{e}_j)^\top, \quad \lambda_{ij} \ge 0$ is the multiplier.  
\end{thm}
\begin{remark}
Note that Theorem \ref{lipsdpthm} is different from Theorem 2 in \cite{fazlyab2019efficient}, as $T$ in our case is a diagonal matrix. This diagonal choice avoids the counterexamples reported in~\cite{pauli2021training} for the full-matrix network relaxation.
% To address tractability, LipSDP offers three variants: LipSDP-Network, LipSDP-Neuron, and LipSDP-Layer formulations.
% which provide a trade-off between computational scalability and estimation accuracy.
  % Among these, LipSDP-Layer and LipSDP-Neuron variants are provably valid; Pauli et al.~\cite{pauli2021training} demonstrate a counterexample where the LipSDP-Network variant fails to yield a strict upper bound. Hence, $T= \sum_{i=1}^{n}\lambda_{ii}e_{i}e_{i}^{\top},\lambda_{ii}\geq 0$ must be restricted to diagonal matrices.
% However, Patricia et. al. \cite {pauli2021training} demonstrates a counter example where LipSDP-Network may fail to produce valid Lipschitz value, rendering only LipSDP-Neuron and LipSDP-Layer valid. Hence, the admissible matrix $T= \sum_{i=1}^{n}\lambda_{ii}e_{i}e_{i}^{\top},\lambda_{ii}\geq 0$ is only diagonal matrix. 
\end{remark}
\begin{figure}
    \centering
    \includegraphics[width=\linewidth]{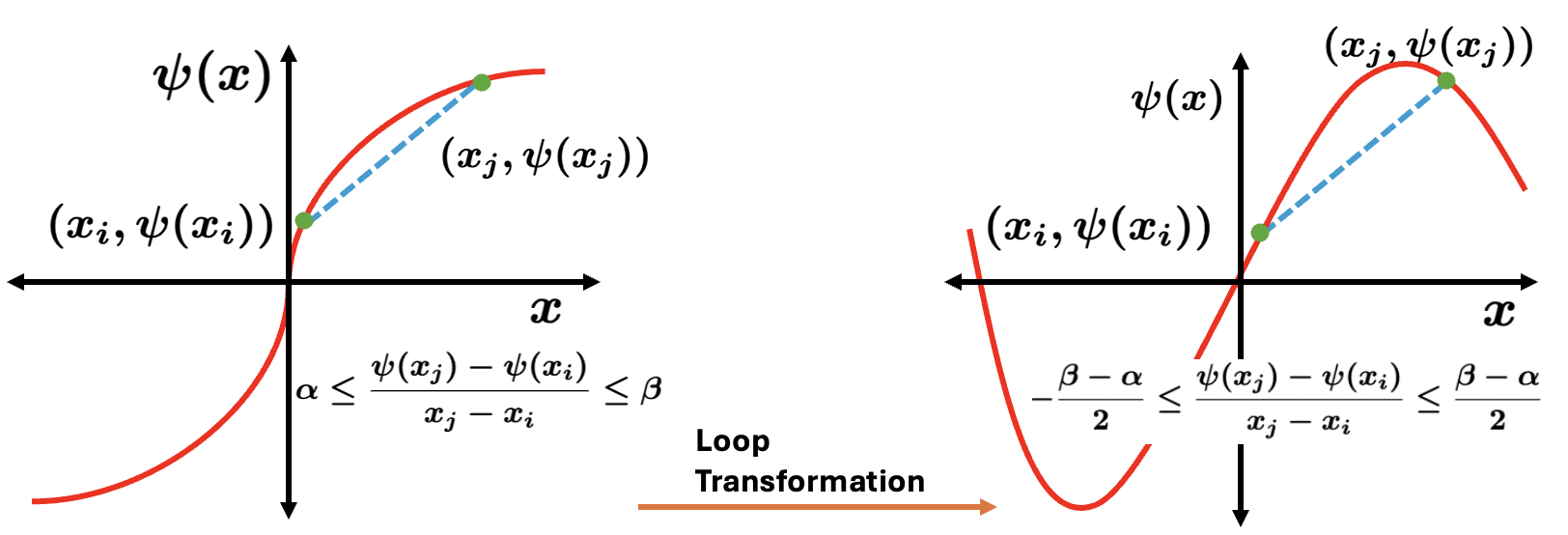}
    % \vspace{1mm} % optional space between figures
    % \includegraphics[width=0.5\linewidth]{looptransformation.png}
    \caption{Illustration of slope bounds for nonlinearity and loop transformation}
    \label{fig:combined_figures}
\end{figure}
Given Theorem~\ref{lipsdpthm}, the tightest certified Lipschitz bound can be computed by solving the following SDP:

% Theorem \eqref{lipsdpthm} provides the sufficient condition for an upper bound on the Lipschitz constant. Thus, the smallest value for Lipschitz bound is determined by solving the following SDP
\begin{equation}\label{eq5}
    \min_{L^{2},T}L^{2},\quad \text{s.t.} \quad \eqref{eq4} \text{~~is satisfied.}
\end{equation}

% The optimization problem \eqref{eq5} can be used to evaluate the robustness of a neural network after the training process. However, in this work, our aim is to learn robust NN by employing these certificates during the training phase, which can be done using \eqref{eq5} to update the weights while minimizing the bound on the Lipschitz constant.
The LipSDP condition \eqref{eq4} is convex in $(T,L^2)$, if the weights matrix N is given.
% , and thus we can efficiently compute the Lipschitz estimate.
However, this condition is computationally intractable for NN synthesis, as it is nonlinear and nonconvex if we search over decision variables $L^2, T$ and $W = (W^0, \dots, W^l)$.

	% \section{Convexified learning under specifications}\label{convexifiedsection}
% In \cite{pauli2021training,wang2023direct}, $\alpha_{\phi}$ is set to zero in \eqref{eq4} and choose a fixed $T$ to formulate the convex constraints by applying the Schur complement to \eqref{eq4}, and rearrange matrix inequality, resulting in an equivalent semidefinite constraint. For instance for a single-layer case constraint \eqref{eq4} reduced to
\section{Convexified learning under specifications}\label{convexifiedsection}
Prior works~\cite{pauli2021training,wang2023direct} derive convex LipSDP constraints by fixing a matrix $T$ and setting $\alpha_\phi = 0$ in~\eqref{eq4}, after that applying the Schur complement to~\eqref{eq4} yields an equivalent semidefinite representation. For example, in a single-layer network, the constraint \eqref{eq4} reduces to:
\begin{align}\label{eq:sdp_single}
\begin{bmatrix}
 -L^{2}I & \beta W^{0\top} T & 0 \\
 \beta T W^{0} & -2T & W^{1\top} \\
 0 & W^{1} & -I
\end{bmatrix} \preceq 0
\end{align}	

Although~\eqref{eq:sdp_single} is convex in $(L,W)$ when $T$ is fixed, it becomes bilinear in $(W^{0},T)$ and hence non-convex if $T$ is also optimized. 
% when jointly optimize over the weights and certificate parameters.
Moreover, choosing fixed $T$ introduces conservatism
% into the framework
and typically requires solving an auxiliary LipSDP problem to estimate $T$ on a vanilla or $\ell_2$-regularized network~\cite{pauli2021training}, which adds an additional layer to computations.
% and requires a suitable choice for $T$. Patricia et. al. \cite{pauli2021training} determine the matrix $T$ from the Lipschitz constant estimation outlined in \cite{fazlyab2019efficient} on the vanilla NN or the L2 regularized NN trained on the same problem. But this approach is computationally expensive, adding an additional layer of solving LipSDP.
Additionally, restriction on $\alpha_{\phi}=0$ forces the activation sector to $[0, \beta_{\phi}]$, omitting the lower slope information that is present in widely-used activations e.g., $\tanh$, and leaky ReLU. This results in a conservative sector approximation that reduces the regularization effect of constraint and weaker bounds on the Lipschitz constant. Moreover, the assumption of nonnegative slope implicitly rules out non-monotonic activations. 
% like Swish or sine, and tends to overestimate slopes in saturated regions of bounded functions like sigmoid and $\tanh$. From a training perspective, this simplification reduces the effectiveness of the regularization: the absence of a lower slope limit weakens the dual variables associated with the constraint, yielding flat gradients that can hinder convergence or lead to suboptimal solutions. Additionally, setting $\alpha_{\phi} = 0$ limits the activation design space by precluding the use of custom or learnable activations whose slopes span both positive and negative values..  
To alleviate those limitation we perform \textit{loop transformation}~\cite{ul2022learning} as shown in Fig~\eqref{fig:combined_figures}, normalizes the nonlinearity $\tilde{\phi}$ to lie in the sector $[-1_{n_{\phi}\times 1},1_{n_{\phi}\times 1}]$, supports general activation structures without restriction on $\alpha_{\phi}, \beta_{\phi}$ or $T$ and enables a convex training with broader applicability.

\subsection{Loop Transformation}\label{looptrasnform}
	The key idea is to reparametrize the NN such that the condition \eqref{eq4} becomes convex in a transformed space. 
    % To streamline the presentation, we derive our results for a two-layer NN with $l=1$ hidden layer, that is $\Psi(x)=W^1\phi(W^0x+b^0)+b^1$. In this case, equation \eqref{nn1} can be rewritten as
	% \begin{gather}\label{1nn}
	% \begin{bmatrix} v^0\\ \Psi(x)  \end{bmatrix}
	% =
	% N\begin{bmatrix} x \\ x^1  \end{bmatrix}+ \begin{bmatrix} b^0 \\ b^1  \end{bmatrix}\\
	% x^1=\phi (v^0),
	% \end{gather}
	% where matrix $N$ depends on the weights as
	% \begin{gather}\label{19}
	% N=
	% \begin{bmatrix}N_{vx} & N_{vx^1} \\ N_{\Psi x} & N_{\Psi x^{1}} \end{bmatrix}=
	% \begin{bmatrix}W^0 & 0 \\ 0 & W^1 \end{bmatrix}.
	% \end{gather}
	Loop transformation is a standard linear fractional transformation technique in  the control literature\textcolor{black}{\cite{yin2021imitation}}. 
    This transformation leads to new representation of the NN (for detailed derivation see our previous work \cite{ul2022learning}):
	\begin{gather}\label{A1}
	\begin{bmatrix} v_{\phi}\\  \textcolor{black}{\Psi(x)} \end{bmatrix}
	=
	\tilde{N}\begin{bmatrix} x \\ \tilde{x}^1  \end{bmatrix}+ \begin{bmatrix} \tilde{b}_{\phi} \\ \tilde{b}^l  \end{bmatrix}\\
	\tilde{x}_{\phi}=\tilde{\phi} (v_{\phi}).
	\end{gather}
	where 
	$\tilde{b}_{\phi}=\left( I-C_2\right) ^{-1} b_{\phi}$, $\tilde{b}^l=C_4\left( I-C_2\right) ^{-1}b_{\phi}+b^l$,  with $C_1=N_{vx^1}\frac{ \beta_\phi -\alpha_\phi}{2},~~ C_2=N_{vx^1}\frac{ \beta_\phi +\alpha_\phi}{2},
	    C_3=N_{\Psi x^{1}}\frac{ \beta_\phi -\alpha_\phi}{2},~~ C_4=N_{\Psi x^{1}}\frac{ \beta_\phi +\alpha_\phi}{2}$, 
    and	\begin{align}\label{27}
	    \begin{split}
	          \tilde{N}&=\begin{bmatrix} \left( I-C_2\right) ^{-1}N_{vx} & \left( I-C_2\right) ^{-1}C_1 \\  N_{\Psi x}+C_4\left( I-C_2\right) ^{-1}N_{vx}& C_3+C_4\left( I-C_2\right) ^{-1}C_1 \end{bmatrix}\\&:=\begin{bmatrix}
	          \textcolor{black}{\tilde{N}_{vx}} &\textcolor{black}{\tilde{N}_{vx^{1}}}\\\textcolor{black}{\tilde{N}_{\Psi x}}&\textcolor{black}{\tilde{N}_{\Psi x^{1}}}
	          \end{bmatrix}
	    \end{split}
	\end{align}
	% It can be seen that $\tilde{N}$ is in general a nonlinear function of~$N$.
    % To solve the equation, an ADMM algorithm is developed in~\cite{yin2021imitation}. 
    Note that $\tilde{N}$ is indirectly depends on $N$ through the bounds $\alpha_{\phi},\beta_{\phi}$.
    %$\left( A_\phi,B_\phi \right) $.
    In particular, assume that the state limits and $N$ are provided. Then $\tilde{N}$ is constructed by: $(i)$ propagating the input bounds
    % \textcolor{black}{the bounds on $x$} 
     through NN to obtain 
     % $\underline{v},\bar{v}$ 
     pre-activation input $v^k$ bounds, $(ii)$ calculate local slope bounds $\alpha_{\phi},\beta_{\phi}$
    % $A_{\phi},B_{\phi}$
    consistent with the activation, and $(iii)$ 
    % performing steps to 
    computing $\tilde{N}$ from $(N, \alpha_{\phi},\beta_{\phi})$. Hereafter we treat $\tilde{N}$ as decision variable instead of $N$ (i.e., $\tilde{N}=f(N)$ is reparametrization of $N$).

	\subsection{Admissibility Condition After Loop Transformation}
	% Now we analyze the admissibility of NN after loop transformation. Based on the new representation of NN, the new matrix inequality in \eqref{eq4} can be written as
    Based on the transformed representation of NN, the Lipschitz admissibility condition \eqref{eq4} can be written as:
    \begin{align}\label{lmi21}
\begin{split}
     \begin{bmatrix} \tilde{N}_{vx} & \tilde{N}_{vx_{\phi}} \\ 0 &I_{n_{\phi}} \end{bmatrix}^{\top} &
  \begin{bmatrix}  T & 0 \\ 0 & -T \end{bmatrix}  \begin{bmatrix} \tilde{N}_{vx} & \tilde{N}_{vx_{\phi}} \\ 0 &I_{n_{\phi}} \end{bmatrix}\\&+
		\begin{bmatrix} \tilde{N}_{\Psi x}^{\top}\tilde{N}_{\Psi x}-L^{2}I & \tilde{N}_{\Psi x}^{\top}\tilde{N}_{\Psi x_{\phi}} \\ \tilde{N}_{\Psi x_{\phi}}^{\top}\tilde{N}_{\Psi x} & \tilde{N}_{\Psi x_{\phi}}^{\top}\tilde{N}_{\Psi x_{\phi}} \end{bmatrix}  \preceq 0
\end{split}
  \end{align}
  % The new LipSDP condition \eqref{lmi21} is convex in $T$ and $L^2$ with given $NN$. However, the
  Since our main goal is to train the NN, 
  and treat $\tilde{N}$ as a decision variable 
  along with $T$ and $L^2$. 
  For that we try to derive the condition that is convex in all those variables. After solving \eqref{lmi21}, we have
  % the first term of \eqref{lmi21}, we have
%   \begin{align}\label{lmi212}
% \begin{split}
%      &\begin{bmatrix} \tilde{N}_{vx}^{\top} T\tilde{N}_{vx} &\tilde{N}_{vx}^{\top} T\tilde{N}_{vx_{\phi}}\\ \tilde{N}_{vx_{\phi}}^{\top}T\tilde{N}_{vx} & \tilde{N}_{vx_{\phi}}^{\top}T\tilde{N}_{vx_{\phi}}-T \end{bmatrix}\\&+
% 		\begin{bmatrix} \tilde{N}_{\Psi x}^{\top}\tilde{N}_{\Psi x}-L^{2}I & \tilde{N}_{\Psi x}^{\top}\tilde{N}_{\Psi x_{\phi}} \\ \tilde{N}_{\Psi x_{\phi}}^{\top}\tilde{N}_{\Psi x} & \tilde{N}_{\Psi x_{\phi}}^{\top}\tilde{N}_{\Psi x_{\phi}} \end{bmatrix}  \preceq 0
% \end{split}
%   \end{align}

  % Condition \eqref{lmi212} can further be written as
%   \begin{align}\label{lmi213}
% \begin{split}
%      &\begin{bmatrix} \tilde{N}_{vx}^{\top}  \\ \tilde{N}_{vx_{\phi}}^{\top} \end{bmatrix}\begin{bmatrix}
%          T
%      \end{bmatrix} \begin{bmatrix}
%          \tilde{N}_{vx}&\tilde{N}_{vx_\phi}
%      \end{bmatrix}-\begin{bmatrix}
%          L^{2}I&0\\0&T
%      \end{bmatrix}\\&+
% 		\begin{bmatrix} \tilde{N}_{\Psi x}^{\top} \\ \tilde{N}_{\Psi x_{\phi}}^{\top} \end{bmatrix} \begin{bmatrix}
% 		    I
% 		\end{bmatrix}\begin{bmatrix}
% 		    \tilde{N}_{\Psi x}&\tilde{N}_{\Psi x_{\phi}}
% 		\end{bmatrix} \preceq 0
% \end{split}
%   \end{align}

  \begin{align}\label{lmi215}
\begin{split}
     \begin{bmatrix} \tilde{N}_{vx}^{\top} & \tilde{N}_{\Psi x}^{\top} \\ \tilde{N}_{vx_{\phi}}^{\top} &\tilde{N}_{\Psi x_{\phi}}^{\top} \end{bmatrix} &
  \begin{bmatrix}  T & 0 \\ 0 & I \end{bmatrix} \begin{bmatrix}
      \star
  \end{bmatrix} 
  % \begin{bmatrix} \tilde{N}_{vx} & \tilde{N}_{vx_{\phi}} \\ \tilde{N}_{\Psi x} &\tilde{N}_{\Psi x_{\phi}} \end{bmatrix}\\&
  -
		\begin{bmatrix} L^{2}I & 0 \\ 0 & T \end{bmatrix}  \preceq 0 
\end{split}
  \end{align}
  Applying Schur compliments yields an equivalent condition 
  \begin{align}\label{lmi216}
      \begin{bmatrix}
          L^{2}I &0 & \tilde{N}_{vx}^{\top} &\tilde{N}_{\Psi x}^{\top} \\ 0&T&\tilde{N}_{v x_{\phi}}^{\top} &\tilde{N}_{\Psi x_{\phi}}^{\top}\\ \tilde{N}_{vx} &\tilde{N}_{vx_{\phi}} &T^{-1}&0\\ \tilde{N}_{\Psi x} &\tilde{N}_{\Psi x_{\phi}} &0 &I
      \end{bmatrix} \succeq 0
  \end{align}
  The condition \eqref{lmi216} is linear in $\tilde{N}$, but still non-convex in $T$. Multiplying \eqref{lmi216} on the left and right by $diag\left( \left[\begin{smallmatrix}
	I_{n_x}&0\\0&\textcolor{black}{T}^{-1}
	\end{smallmatrix}\right],I_{n_{\phi}},I_{n_{y}}\right)$, we obtain

\begin{align}\label{finallmi}
    LMI(Q,L,K):=\begin{bmatrix}
          L^{2}I &0 & K_{1}^{\top} &K_{3}^{\top} \\ 0&Q_{1}&K_{2}^{\top} &K_{4}^{\top}\\ K_{1} &K_{2} &Q_{1}&0\\ K_{3} &K_{4} &0 &I
     \end{bmatrix} \succeq 0
  \end{align}
  where $Q_{1}=T^{-1}\succ 0$, $K_{1}=\tilde{N}_{vx}$, $K_{2}=\tilde{N}_{vx_{\phi}}Q_{1}$, $K_{3}=\tilde{N}_{\Psi x}$ and $K_{4}=\tilde{N}_{\Psi x_{\phi}}Q_{1}$. Now the condition \eqref{finallmi} is convex in the decision variables $\left(Q_{1},K_{1},K_{2},K_{3},K_{4}\right)$. Variable $\left(L,T,\tilde{N}\right)$ that satisfy the condition \eqref{lmi21} can be recovered using the computed $\left(Q_{1},K_{1},K_{2},K_{3},K_{4}\right)$ through the following equations: $T=Q_{1}^{-1}$ and $\tilde{N}=KQ^{-1}$, where 
  \begin{align}
      Q:=\begin{bmatrix}I &0\\0&Q_{1}
     \end{bmatrix},\quad \text{and} \quad K:=\begin{bmatrix}
         K_{1} &K_{2}\\K_{3} &K_{4}
     \end{bmatrix}
  \end{align}
  Thus the convex condition \eqref{finallmi} allow us to search over $L,T,$ and $\tilde{N}$ simultaneously.
 \subsection{Algorithm}
 % In general, NN are trained on by minimizing task-specific loss, such as mean-squared error or cross-entropy.
 In our study, we aim to minimize not only network loss, but also certified Lipschitz bounds, resulting in a constrained optimization problem with separable objectives:
  \begin{equation}\label{mainproblem}
 \begin{split}
\min_{N,Q,L,K}\mathcal{L}(N)+ \eta L^{2}\\
\text{s.t}~~ LMI(Q,L,K) \succeq 0\\
f(N)Q=K
 \end{split}
 \end{equation}
 The loss $\mathcal{L}(N)$ is an explicit function of network parameters and the Lipschitz bound $L$ depends on $\tilde{N}$ via LMI~\eqref{finallmi} and $\eta>0$ 
 % is a weighting parameter adjusting
control the trade-off between accuracy and the robustness. The optimization includes two sets of decision variables: $N$ and $(Q,L,K)$. To solve~\eqref{mainproblem}, we use Alternating-Direction Method of Multipliers (ADMM), which decomposes problem into smaller subproblems 
 % that are easier to handle individually 
 and solves the constrained problem through alternating minimization steps on the equivalent augmented Lagrangian of the optimization problem \eqref{mainproblem} and the dual update step. We fist define the augmented loss function
 % In order to apply the ADMM algorithm, the objective must be separable. The objective at hand, i,e., the NN loss and the Lipschitz bound are indeed separable, yielding the following optimization problem 
 % In general, NN are trained on input-output data with the objective is of minimizing a predefined loss, e.g., the mean square error, cross-entropy, or hinge loss. We propose to not only minimize the NN's loss but also its Lipschitz constant. This yield an optimization problem with two separate objectives that can be solved conveniently using alternating direction method of multipliers (ADMM). ADMM is an algorithm that solves the optimization problem by splitting them into smaller subproblems that are easier to handle individually. In order to apply the ADMM algorithm, the objective must be separable. The objective at hand, i,e., the NN loss and the Lipschitz bound are indeed separable, yielding the following optimization problem 
%  \begin{equation}\label{mainproblem}
%  \begin{split}
% \min_{N,Q,K}\mathcal{L}(N)+ \eta L^{2}\\
% \text{s.t}~~ LMI(Q,L,K) \succeq 0\\
% f(N)Q=K
%  \end{split}
%  \end{equation}

 \begin{equation}
 \begin{split}
    \mathcal{L}_{\rho}&(N,Q,L,K,Y) =\mathcal{L}(N)+\eta L^{2}\\&+ \text{tr}\left(Y^{\top}\cdot(f(N)Q-K)\right)+\frac{\rho}{2}\|f(N)Q-L\|^{2}_{F}
 \end{split}
 \end{equation}
 where $\|\cdot\|_{F}$ is the Frobenius norm, $Y$ is the Lagrange multiplier and $\rho>0$ is penalty parameter that influence the convergence behavior of ADMM. The solution for~\eqref{mainproblem} is then determined via following iterative ADMM update steps:
\begin{equation}\label{nupdate}
N^{i+1}=\arg\min_{N}\mathcal{L}_{\rho}(N,Q^{i},L^{i},K^{i},Y^{i})
\end{equation} 
\begin{align}\label{sdpupdate}
(L,Q,K)^{i+1} &= \arg\min_{L,Q,K}\; \mathcal{L}_{\rho}(N^{i+1}, Q, L, K, Y^i) \nonumber \\
 \quad & \text{s.t.}\quad\text{LMI}(Q,L,K) \succeq 0
\end{align}

% \begin{equation}\label{sdpupdate}
% \begin{split}
% \left(L,Q,K\right)^{i+1}=\arg\min_{L,Q,K}\mathcal{L}_{\rho}(N^{i+1},Q,L,K,Y^{i})\\ \text{s.t}\quad LMI(Q,K,L)\succeq 0
% \end{split}
% \end{equation}
If $\|f(N^{i+1})Q^{i+1}-K^{i+1}\|_{F}\leq \sigma$, where $\sigma>0$ is predefined tolerance, then the algorithm has converged, and 
% we have found the solution to \eqref{mainproblem}, so 
terminate. Otherwise update $Y$ and return to $N$-update step \eqref{nupdate}.
\begin{equation}\label{yupdate}
    Y^{i+1}=Y^{i}+\rho \left(f(N^{i+1})Q-K^{i+1}\right)
\end{equation}
 
The unconstrained update in \eqref{nupdate} can be solved using a gradient-based algorithm, while \eqref{sdpupdate} is a convex SDP handled by SDP solver. 
% while the convex subproblem in~\eqref{sdpupdate} is handled effectively using SDP solvers. 
The dual variable $Y$ in \eqref{yupdate} accumulates violations of the constraint $f(N)Q=K$, enforcing consistency over iterations. Each step is well-posed: gradient descent in $N$-update typically converges to local optima, and $(L,Q,K)$-update in \eqref{sdpupdate} always obtains global optimum. However, since $\mathcal{L}(N)$ and constraint $f(N)Q = K$ are nonconvex, global convergence of the overall ADMM procedure is not guaranteed. Theoretical understanding of ADMM in nonconvex settings remains an active area of research.
% However, the loss $\mathcal{L}(N)$ and the constraint $f(N)Q=K$ are nonconvex, and hence the proposed ADMM does not guarantee convergence to a global optima. Convergence properties of ADMM without convexity are still subject to ongoing research. 
Nonetheless, any converged solution provides a robust NN with safety guarantees.  

% \begin{remark}
% % The computational cost of the framework is dominated by the semidefinite program in the $(L, Q, K)$ update step. The number of decision variables grows with the number of activation functions, which is in turn determined by the number of layers and neurons per layer. Since SDP solvers typically scale cubically with problem size, this step can become a computational bottleneck for networks with high-dimensional inputs or deep architectures. For example, in image classification tasks with large input dimensions, the resulting SDP may become intractable without further structural simplifications.
%     Computational scalability and complexity of the proposed framework depends on the number of decision variables of the SDP. The complexity primal/dual SDP solvers scales cubically with the number of decision variables. Hence, the Lipschitz update step  becomes the bottleneck of the proposed method 
%     % as the number of neurons per hidden layersand the number of hidden layers, that together determine thesize of the weights,increases 
%     i.e., $(L,Q,K)$-update will be computationally expensive if the number of activation function $n_{\phi}$ is large. 
%     % For example for picture inputsas commonly used in classification problems, the input dimension is usually high, potentially leading to high computationtimes or computational intractability.
% \end{remark} 
\begin{remark}
The computational bottleneck lies in the SDP step \eqref{sdpupdate}, which scales cubically with the number of activation functions $n_{\phi}$. As $n_{\phi}$ increases with network width and depth, scalability to large networks may require further approximation or decomposition.
\end{remark}

\section{Scalable Lipschitz Certification via RS-LMI}
In Section~\ref{convexifiedsection}, we presented a SDP-based framework to enforce global Lipschitz constraints during training. However, solving semidefinite constraints in \eqref{sdpupdate} is computationally expensive
% non-trivial to deal with
for large-scale problem. One direction for improving scalability is to move the semidefinite constraints into the cost function via the exact penalty method~\cite{wang2024scalability}. Although this approach makes the SDP differentiable compatible with TensorFlow auto-diff tool, but it still requires $\mathcal O\bigl((\sum n_k)^3\bigr)$ eigenvalue decomposition over $(\sum_k n_k)\times (\sum_k n_k)$ matrix, making per-batch training memory intensive.
% The global LipSDP constructs a single large LMI over all layers, whose optimal value $L$ yields the tightest Lipschitz constant.  However, solving a global $(\sum_k n_k)^2$-dimensional LMI incurs a computational complexity of $\mathcal O\bigl((\sum n_k)^3\bigr)$ and requires external SDP solvers and ADMM loops, making it impractical for even moderate-sized networks. Currently, general purpose SDP solvers are not capable of addressing the scalability issue of LipSDP, placing a significant hurdle for practical use of such methods. Xu et. al. \cite{chord2, chord3} exploit the chordal sparsity contained in SDPs. Chordal sparsity allows us to decompose a large SDP constraints to set of smaller and coupled ones which might be more scalable to solve.  However, these methods still cannot scale to practical networks on CIFAR10 or ImageNet. In this section, we present RS-LMI, a novel framework that (i) decomposes the global certificate into per‐layer constraints, (ii) sketches each constraint to a tiny subspace, and (iii) incorporates them as a smooth, auto differentiable penalty.
% To improve the memory efficiency and scalability, we propose \textit{RS-LMI}, that replace big SDP with per layer LMIs.

We begin by observing that global Lipschitz certificate can be decomposed into independent per-layer conditions under mild assumptions:
\begin{thm}\label{spec}
Assume each activation $\phi$ is 1-Lipschitz, then global LipSDP can split into $l$ independent layer-wise SDPs:
\begin{equation}\label{rslmiconstraint}
 \min_{\tau_{k}\geq 0}\tau_{k}~~\text{s.t}~~  \begin{bmatrix}
   \tau_{k} I_{n_k} & W_k^T\\ W_k& I\end{bmatrix}\succeq 0 \Leftrightarrow \tau_{k}I-W_{k}^{\top}W_{k}\succeq 0
\end{equation}
and $\tau^{*}_{k}=\sigma_{\max}(W_{k})^2$ is unique optimal solution to SDP~\eqref{rslmiconstraint} for each layer $k$, where $\sigma_{max}$ denote the singular value. If $L$ be the Lipschitz constant for global LipSDP, then $L=\sqrt{\tau^{*}_{1}\cdots\tau^{*}_{l}}=\prod_{k=1}^{l}\sqrt{\tau^{*}_{k}}=\prod_{k=1}^{l}\sigma_{\max}(W_{k})$.
\end{thm}
This decomposition avoids solving a large joint SDP, reducing both memory and runtime costs. However, solving each per-layer SDP still incurs $\mathcal{O}(n_k^3)$ complexity per layer, which becomes costly for wide networks.
% Theorem \ref{spec} highlight under 1-Lipschitz activation the product of layer spectral-norms is exactly as tight as the LipSDP. Although this per-layer SDP reduces the problem size, but still cost $\mathcal{O}(n_{k}^{3})$. 
% For wide networks, these computation remain a bottleneck.
To reduce this cost, we leverage the idea of randomized sketching. Formally, for each layer $k$, we draw a fixed Gaussian sketch matrix $G_k \in \mathbb{R}^{n_{k-1} \times m}$ with~$m \ll n_{k-1}$, and replace the constraints with the sketched LMI:
\[
G_k^\top (\tau_k I - W_k^\top W_k) G_k \succeq 0.
\] 
This projection reduces the computational cost of each constraint to $\mathcal{O}(n_k m^2 + m^3)$, while preserving Lipschitz bound with high probability, as guaranteed by Johnson–Lindenstrauss lemma \cite{pilanci2015randomized}. Nevertheless, solving the resulting projected SDPs at each iteration still requires external solvers. To bypass this bottleneck, we instead encode each sketched constraint as a differentiable penalty term:
% However, directly solving the resulting SDP for each mini-batch remains memory intensive and require heavy solver. Therefore, to this end, we reformulate the sketched LMI as a differentiable penalty function. This allows the certification to be integrated directly into training via gradient-based optimization, without requiring external solvers. Specifically, we define the penalty:
\[
\mathcal{P}_k(W_k, \tau_k) = \left\| \left[ G_k^\top W_k^\top W_k G_k - \tau_k I \right]_+ \right\|_F^2,
\]
where $[\cdot]_+$ denotes the projection onto the positive semidefinite cone by zeroing negative eigenvalues. 
% This penalty quantifies the violation of the sketched constraint, and vanishes when the constraint is satisfied. 
By incorporating these penalties into the loss function, we obtain the following optimization problem
\begin{eqnarray}
   \min_{N,\tau_{k},\alpha_k}  \mathcal{L}(N) + \sum_{k=1}^L \left[ \tau_k + \alpha_k \mathcal{P}_k(W_k, \tau_k) \right],
\end{eqnarray}
where 
% $\mathcal{L}(N)$ is the task-specific loss, and
$\alpha_k > 0$ is a penalty weight that balances robustness and accuracy. This formulation is fully differentiable and compatible with standard gradient-based optimizers. 
% Unlike norm-clipping methods, it allows each $\tau_k$ to be trained alongside $W_k$, enabling adaptive allocation of the Lipschitz budget across layers.
 % and enabling seamless integration in PyTorch. 
 % Unlike fixed-margin methods, we treat each certificate $\tau_k$ as a trainable parameter, jointly optimized with weights $W_k$, allowing the model to adaptively allocate its Lipschitz budget. 

\begin{remark}
    RS-LMI reduces the per-batch complexity to $\sum_k \mathcal{O}(n_k m^2 + m^3)$, offering $10$–$50\times$ speedup over per-layer SDPs with negligible loss in bound quality. At convergence, $\|W_k\|_2 \le \sqrt{\tau_k}$ with high probability, and $\prod_{k=1}^{l} \sqrt{\tau_k}$ provides a valid global certificate. Unlike spectral-norm clipping, RS-LMI captures multi-directional variations in weight matrices and avoids infeasibility or high memory usage.
    % RS-LMI captures multi-directional interactions and supports smooth optimization, unlike spectral-norm clamping. It avoids infeasible solves and high memory of global SDPs, and replaces cubic-cost eigen-solves in per-layer SDPs with efficient $m$-dimensional sketches, yielding near exact probabilistic certification at low cost.
\end{remark}

\begin{figure}[t]
    \centering
    \subfloat[Test-time prediction on 2D synthetic data.\label{fig1}]{
        \includegraphics[width=0.43\linewidth]{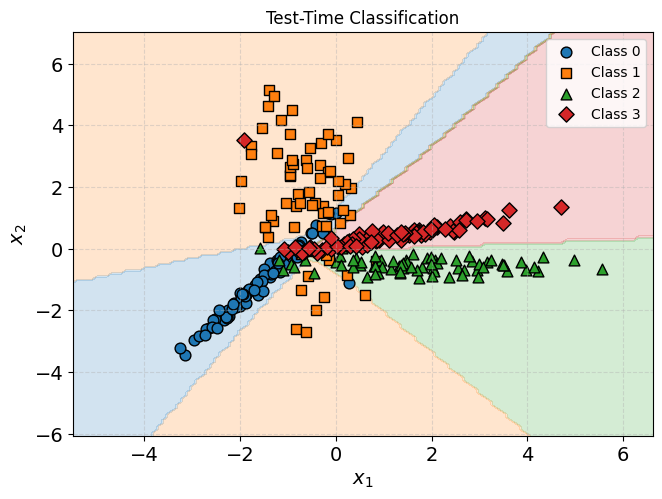}
    }
    \hfil
    \subfloat[Accuracy of Lip-Loop on noise-corrupted MNIST test data.\label{fig2}]{
        \includegraphics[width=0.5\linewidth]{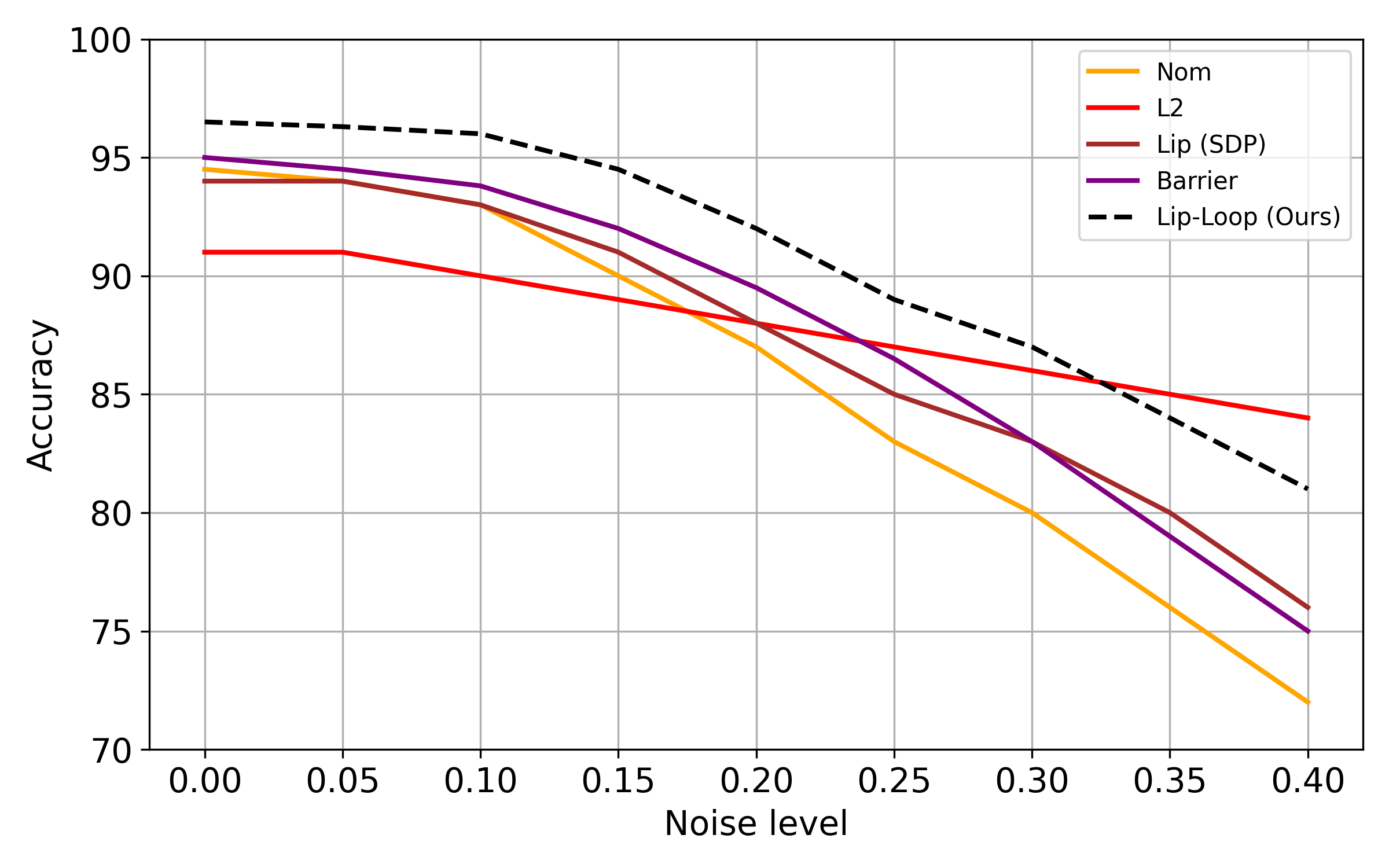}
    }
    \caption{Performance of Lip-Loop under distribution shifts.}
    \label{fig:results}
\end{figure}

 \section{Experimental Results}
 
We evaluate the proposed algorithms on four benchmarks: a synthetic 2D classification task, MNIST, CIFAR-10 and Image-Net to access 
% their effectiveness in 
certifiable robustness, predictive accuracy and computational efficiency. We aims to answer the following questions: \textbf{(RQ1)} How do our methods compare to state-of-the-art techniques in robustness and accuracy? \textbf{(RQ2)} Whether \textit{RS-LMI} recover the same bound as \textit{Lip-Loop} when both problems are fully solved? \textbf{(RQ3)} Does \textit{RS-LMI} improves computational efficiency and reduce memory usage? \textbf{(RQ4)} How well does RS-LMI scale to larger networks and datasets?
% Can RS-LMI scalable to large networks and generalize across different model sizes?

 For all experiments we compare against several baselines: \textbf{Nom-NN} denotes standard NN training without constraints.
 % feedforward neural network trained without robustness constraints. 
 \textbf{$\boldsymbol{\ell_2}$-NN} incorporates weight decay constraints via $\ell_2$ regularization. \textbf{Lip-NN} solves the Lipschitz-constrained SDP 
% using ADMM approach 
suggested by \cite{pauli2021training}. \textbf{Barrier method} solves bilinear SDP using log-determinant barrier \cite{pauli2022neural}. For larger-scale models, we include \textbf{GloRo}, which estimates the Lipschitz constant via spectral norm product heuristic \cite{leino2021globally}.
% commonly used in high-dimensional setting such as ImageNet \cite{leino2021globally}. We also compare with 
\textbf{LipDiff}, reformulates the SDP as an eigenvalue optimization problem compatible with autodiff frameworks, and \textbf{LipDiff-Ex}, which uses exact eigenvalues instead of Lanczos approximations \cite{wang2024scalability}. 
% Our proposed method \textbf{Lip-Loop}, convexify the constraints using loop transformation and optimizes it using ADMM, while \textbf{RS-LMI}, replaces the global SDP constraints with layerwise randomized subspace LMIs that are differentiable and computationally efficient.

To answer the \textbf{RQ1 \& RQ2}, Table \eqref{tab:robustness_results} summarizes the certified Lipschitz bounds, test accuracy, and cross-entropy loss (CEL) on the 2D and MNIST datasets. We observe that Lip-Loop achieves a significantly lower Lipschitz constant than all baseline while maintaining high test accuracy,
indicating the benefit of its reparameterized training formulation. On the 2D task, Lip-Loop improves the certified bound by 80\% compared to Nom-NN, while achieving 91.2\% accuracy. The Barrier method performs competitively but shows lower accuracy, likely due to challenges in optimizing non-convex bilinear SDPs. 
% $\ell_2$-NN yields moderate robustness but suffers from underfitting, while Lip-NN provides improved robustness over Nom-NN but remains less effective than Lip-Loop due to its reliance on fixed slope bounds.
\textit{RS-LMI} closely matches this results with $L=36.9$ and 90.5\% accuracy. On MNIST, Lip-Loop achieves $L=9.8$ with 97.1\% accuracy, outperforming all baselines, and RS-LMI again align its performance (\(L=10.3\), 97.8\%) with \textit{Lip-Loop}, 
verifying that the randomized subspace relaxation preserves certified bound tightness without sacrificing accuracy.
% Compared to $\ell_2$-NN and Lip-NN, Lip-Loop yields both lower cross-entropy loss and better robustness, indicating the benefit of its reparameterized training formulation.

% On MNIST, Lip-Loop achieves $L=9.8$ with 97.1\% test accuracy, outperforming all baselines in the robustness and accuracy trade-off and RS-LMI again mirror this performance (\(L=10.3\), 97.8\%). 
% The inclusion of RS-LMI as the final row in each dataset serves as a sanity check: its close alignment with Lip-Loop verifies that the sketch-based formulation faithfully approximates the full SDP certificate when fully solved.

% achieves a favorable trade-off, with a certified bound of 9.8 and test accuracy of 97.1\%. These results affirm that convexified training can enforce certified robustness without compromising predictive performance. 
Figure~\ref{fig1} illustrates the test-time behavior of the model trained by \textit{Lip-Loop} on the 2D task, the decision boundaries learned are smooth and well-aligned with the class geometry, suggesting stable gradient behavior across the input domain.
% On MNIST, the predictions from Lip-Loop remain consistent even for visually ambiguous digits, indicating better generalization under uncertainty. 
% These qualitative results support the quantitative trends and illustrate how convex training formulations can encourage global input-output stability.

% Regularization via $\ell_2$ penalties reduces the bound but at the cost of underfitting. Lip-NN and the Barrier method provide improved robustness over nominal models but do not match Lip-Loop in either bound tightness or prediction accuracy. These differences may be attributed to the conservative nature of slope bounds in Lip-NN and the non-convexity inherent in the Barrier formulation. In contrast, Lip-Loop optimizes within a reparameterized convex subspace, which enables more direct control over the function’s input-output sensitivity. On MNIST, a similar pattern is observed. Lip-Loop achieves a Lipschitz bound of 9.8 while maintaining 97.1\% accuracy, outperforming regularized and SDP-based baselines on both robustness and predictive metrics.
%%%%%%%%%%%%%%%%%%%%%%%%%%%%%%%%%%%%%%%%%

\begin{table}[h]
\centering
\caption{CEL on the training data, test accuracy, and certified $L$ on the 2D and MNIST benchmark}
\begin{tabular}{|c|l|c|c|c|}
\hline
Dataset & Model & CEL & Accuracy & $L$ \\
\hline
\hline
{2D} 
  & Nom-NN        & 0.15 & 96.0\% & 339.1 \\
  & $\ell_2$-NN         & 0.42 & 80.5\% & 56.7  \\
  & Lip-NN        & 0.37 & 88.7\% & 49.4  \\
  &Barrier (bilinear SDP) & 0.35 &90.4\% &42.8\\
  & Lip-Loop (Ours) & 0.35 & 91.2\% & 35.2 \\ & RS-LMI (Ours) & 0.36 & 90.5\% & 36.9 \\
\hline
{MNIST} 
  & Nom-NN        & 0.10 & 96.5\% & 140.6 \\
  & $\ell_{2}$-NN         & 0.27 & 90.4\% & 15.3  \\
  & Lip-NN        & 0.21 & 96.3\% & 13.2  \\
  & Barrier (bilinear SDP) &0.19 & 97.2\% & 16.9\\
  & Lip-Loop (Ours) & 0.18 & 97.1\% & 9.8 \\ & RS-LMI (Ours) & 0.20 & 97.8\% & 10.3  \\
\hline
\end{tabular}
\label{tab:robustness_results}
\end{table}

To further probe \textbf{RQ1}, we examine robustness under distributional shift by corrupting MNIST with Gaussian noise $\mathcal{N}(0,\sigma^2)$ and averaging accuracy over 10 trials shown in Fig.~\ref{fig2}. Lip-Loop exhibits the highest accuracy across noise levels, maintaining stable performance under increasing perturbation magnitude, indicating robustness beyond training distribution. In contrast, $\ell_2$-NN exhibits a smoother degradation but starts from a lower accuracy,
% shows the evaluation on noise corrupted data, that was created by adding Gaussian noise $\mathcal{N}(0,\sigma^2)$ to the normalized MNIST data and average over 10 different noise samples. 
% Lip-Loop maintains the highest accuracy across the entire noise range, indicating robustness beyond the training distribution. 
% Interestingly, $\ell_2$-NN exhibits a smoother degradation curve than the other baselines,
% . This behavior
likely
% results 
from the implicit smoothing effect of weight decay, which reduces model sensitivity to noise at the cost of underfitting.
% In contrast, Lip-Loop achieves both high accuracy and controlled degradation by enforcing global Lipschitz constraints throughout training, indicating robustness beyond training distribution.

% Although RS-LMI is designed for scalability in large networks, evaluating it on the same small models allows us to determine whether the randomized eigenvalue-based formulation approximates the original Lip-Loop objective.  The final row in each section of Table \eqref{tab:robustness_results} includes RS-LMI used here as a sanity check. On both dataset, RS-LMI closely matches Lip-Loop in Lipschitz bound and accuracy, confirming that the subspace relaxation preserves robustness when both problems are fully solved.

Finally for \textbf{RQ3 \& RQ4}, we access scalability by comparing computational efficiency and memory usage of the CNN architecture for CIFAR-10 and ImageNet illustrated in Table \eqref{tab:lip_bound_singlecol}. RS-LMI consistently outperforms baselines in efficiency while maintaining competitive accuracy and tight Lipschitz bounds. Training time is reduced by over 90\% on MNIST by GloRo and over 80\% on CIFAR-10 by LipDiff. On Imagnet, RS-LMI is nearly $19\times$ faster than LipDiff. Memory usage improves substantially as RS-LMI uses 96\% less memory than LipDiff on LipDiff and avoid the infeasibility issues encountered by LipDiff-Ex, which fails to run on ImageNet due to large SDP matrices, we are unable to get exact eigenvalue. Bound quality improves by up to 62\% over GloRo. For instance, on CIFAR-10, RS-LMI tightens the bound from 35.45 to 13.58. The bound remains competitive with LipDiff, but at dramatically lower cost.

\begin{table}[!t]
\caption{ RS-LMI achieves the tightest bound while reducing run-time and memory by one–two orders of magnitude.
% LipSDP is out-of-memonry (OOM) on these networks.
% We present result compare the benchmark values of LipSDP, GloRo, LipDiff and its variant. 
% For MNIST-CNN, the SDP constraint size is of $4021\times 4021$ and the SDP solver used for LipSDP triggers the out-of-memory (OOM) error. Furthermore LipDiff-Ex can reduce the Lipschitz constant estimation by $48\%$ compared to norm product method. For CIFAR10-CNN, RS-LMI can reduce the Lipschitz constant by $60\%$ compared to the norm product.
}
\centering
\scriptsize
\setlength{\tabcolsep}{2.3pt} % Reduce column spacing
\begin{tabular}{|c|c|c|c|c|c|c|}
\hline
\textbf{Data} & \textbf{Arch.} & \textbf{} & \textbf{GloRo} & \textbf{LipDiff} & \textbf{LipDiff-Ex} & \textbf{Ours} \\
\hline
\hline
\multirow{4}{*}{MNIST} & \multirow{4}{*}{1C2FC}
& Accuracy & 99\% &98\%  & 97.3\% &  98.8\% \\
& & Lipschitz &25.97 & 14.76 & 13.08  & 15.09 \\
& & Time(s)& 450 & 179 & 559.08  & 39.22\\
& & Memory(MB) & 193.8& 2640 & 1534  & 214.15 \\
\hline
\multirow{4}{*}{CIFAR10} & \multirow{4}{*}{3C3FC}
& Accuracy & 77\% &80\%  & 78\% & 83.13\% \\
& & Lipschitz &35.45 & 14.82 & 18.52 &  13.58\\
& & Time(s) &  5520&  2777.07& 36000  &1101.9 \\
& & Memory(GB) & 0.51& 60.05 & 51.39  & 0.24 \\
\hline
\multirow{4}{*}{ImageNet} & \multirow{4}{*}{ResNet18}
& Accuracy & 35.5\% &66\%  &  infeasible &  63\%\\
& & Lipschitz &195633 &  52546.2&   --- &65464.3  \\
& & Time(s) & 2157.13 & 29647.9 &   ---& 1533.02\\
& & Memory (GB) & 5.28&116.8    & ---& 5.41 \\
\hline
\end{tabular}
\label{tab:lip_bound_singlecol}
\end{table}

	\section{Conclusion and Future Directions}
This letter addresses the challenge of certifiably robust neural network training by enforcing global Lipschitz constraints. We propose Lip-Loop, a loop-transformed convex formulation enabling semidefinite admissibility during training, and RS-LMI, a scalable randomized subspace method for layerwise certification. Our methods yield tighter robustness bounds than prior approaches while reducing training time and memory overhead by an order of magnitude. Future extensions include integrating Lipschitz-based certification into reinforcement learning and adapting the approach for safe decision-making in complex systems.
% We have presented two complementary techniques for certifiable Lipschitz‐constrained learning. Lip-Loop convexifies the classical LipSDP certificate via a loop transformation and integrates it into training with ADMM, yielding tight robustness–accuracy trade‐offs. RS-LMI further scales certification by replacing the global SDP with randomized subspace LMIs, matching full‐SDP bounds on small models and extending to large architectures with orders‐of‐magnitude savings in time and memory.
% Future work will explore applying these certification frameworks to RL, where Lipschitz stability can improve policy robustness, and to in-context learners, where controlled Lipschitz value may enhance generalization and safety in few-shot settings.

\balance
\vskip 0.2in
% \bibliographystyle{ieeetr}
% \bibliography{reference}

\end{document}